# Multi-stage Attention ResU-Net for Semantic Segmentation of Fine-Resolution Remote Sensing Images

Rui Li, Shunyi Zheng, Chenxi Duan, Jianlin Su, and Ce Zhang

*Abstract*—The attention mechanism can refine the extracted feature maps and boost the classification performance of the deep network, which has become an essential technique in computer vision and natural language processing. However, the memory and computational costs of the dot-product attention mechanism increase quadratically with the spatio-temporal size of the input. Such growth hinders the usage of attention mechanisms considerably in application scenarios with large-scale inputs. In this Letter, we propose a Linear Attention Mechanism (LAM) to address this issue, which is approximately equivalent to dot-product attention with computational efficiency. Such a design makes the incorporation between attention mechanisms and deep networks much more flexible and versatile. Based on the proposed LAM, we re-factor the skip connections in the raw U-Net and design a Multi-stage Attention ResU-Net (MAResU-Net) for semantic segmentation from fine-resolution remote sensing images. Experiments conducted on the Vaihingen dataset demonstrated the effectiveness and efficiency of our MAResU-Net. Open-source code is available at https://github.com/lironui/Multistage-Attention-ResU-Net.

## I. INTRODUCTION

Attention mechanisms, benefiting from their powerful ability to exploit long-range dependencies of the feature maps and facilitate neural networks to explore global contextual information, are at the research forefront of computer vision (CV) and natural language processing (NLP) [4, 5]. Dot-product attention mechanisms, generating response at each pixel by weighting features in the previous layer, expand the receptive field to the entire input feature maps in one pass. With a strong ability to capture long-range dependencies, dot-product attention mechanisms have been widely used in vision and language processing tasks. For example, dot-product-attention-based Transformer [6] has demonstrated state-of-the-art performance in nearly all tasks in NLP. The non-local module [8], dot-product-based attention for CV, has demonstrated its superiority on a wide range of vision tasks [4].

However, as the memory and computational overhead of the dot-product attention mechanism increases quadratically, along with the spatio-temporal size of the input, it is difficult to model the global dependency on massive inputs, such as large-scale videos, long sequences, or fine-resolution images, thereby remaining an intractable problem. To alleviate the huge computational costs, Child et al. [10] designed sparse factorizations of the attention matrix and reduce the complexity from $O(N^2)$ to $O(N\sqrt{N})$. Using locality sensitive hashing, Kitaev et al. [11] decreased the complexity to $O(N \log N)$. Furthermore, Katharopoulos et al. [12] and Li et al. [13] took self-attention as a linear dot-product of kernel feature maps, and Shen et al. [14] modified the position of softmax functions, resulting in the complexity reduction at $O(N)$.

In this Letter, we not only reduce the complexity of the dot-product attention mechanism to $O(N)$ from a novel facet, but also increase the classification performance of the U-Net by incorporating the proposed attention and ResNet-based backbones. Specifically, we take the ResNet-34 as the encoder and substitute the plain skip connections within the raw U-Net into our attention blocks at multiple stages, which refine the multi-scale feature maps captured by the network.

As an elaborate encoder-decoder architecture, U-Net [1] has demonstrated its great potential in medical image segmentation [15], land cover classification [16], and UAV drone mapping [17]. Generally, the encoder-decoder-based network includes a contracting part, i.e. encoder, that captures traits of the input and generates corresponding feature maps, and an expanding part, i.e. decoder, in which the mask for pixel-wise segmentation is reconstructed. Specifically, feature maps generated by the encoder are comprised of low-level and fine-grained detailed information, while those of the decoder contain high-level and coarse-grained semantic information [15]. The skip connections, which concatenate the low-level and high-level feature maps, have shown to be effective in enhancing the feature extraction capability of encoder-decoder frameworks. Nevertheless, the plain skip connections in the raw U-Net only connect the features without any further process, leading to

This work was supported in part by the National Natural Science Foundations of China (No. 41671452). *(Corresponding author: Rui Li.)*

R. Li and S. Zheng are with School of Remote Sensing and Information Engineering, Wuhan University, Wuhan 430079, China (e-mail: lironui@whu.edu.cn; syzheng@whu.edu.cn).

Jianlin Su is with the Shenzhen Zhuiyi Technology Co., Ltd. (e-mail: bojonesu@wezhuiyi.com)

C. Duan is with the State Key Laboratory of Information Engineering in Surveying, Mapping, and Remote Sensing, Wuhan University, Wuhan 430079, China; chenxiduan@whu.edu.cn (e-mail: chenxduan@whu.edu.cn).

C. Zhang is with Lancaster Environment Centre, Lancaster University, Lancaster LA1 4YQ, United Kingdom (e-mail: c.zhang9@lancaster.ac.uk).

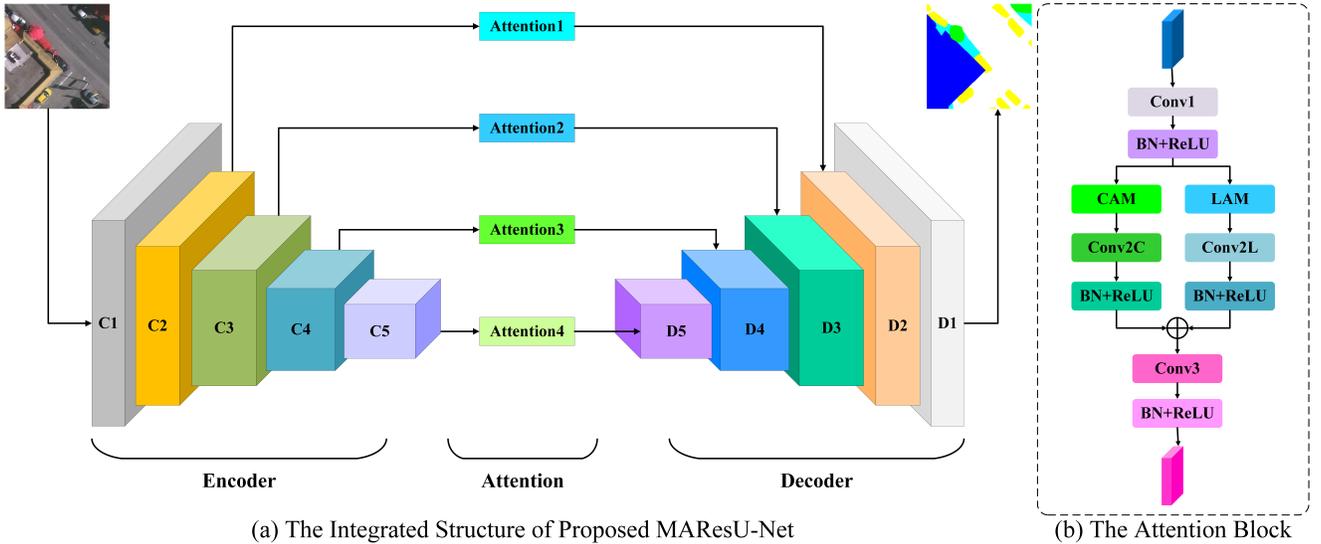

(a) The Integrated Structure of Proposed MAResU-Net  (b) The Attention Block

Fig. 1. The overall architecture of proposed MAResU-Net in (a), and the structure of the attention block in (b).

insufficient utilization of the abundant information. Here, we design a multi-stage attention ResU-Net (MAResU-Net) to remedy this limitation based on the proposed memory-saving and computation-effective linear attention mechanism (LAM). Experiments on Vaihingen semantic segmentation dataset manifest the effectiveness of the proposed network. The major contribution of this Letter includes:

1) We propose a LAM by reducing the complexity of the dot-product-based attention mechanism from $O(N^2)$ into $O(N)$.
2) The proposed LAM allows the combination of attention modules and neural networks that is flexible and versatile.
3) Based on the proposed LAM, we design a multi-stage linear attention U-Net to enhance the performance of the raw U-Net.

## II. RELATED WORKS

### A. Dot-Product Attention

To enhance word alignment in machine translation, Bahdanau et al. [18] proposed the initial formulation of the dot-product attention mechanism. Subsequently, recurrences are completely replaced by attention in Transformer [6]. The state-of-the-art records in almost all tasks in NLP demonstrate the superiority of the attention mechanism. Wang et al. [8] modified the dot-product attention for CV and proposed the non-local module. Other related work conducted on different tasks of CV further showed the effectiveness and general utility of the attention mechanism.

### B. Scaling Attention

Besides dot-product attention, there is another genre of techniques referred to as attention in the literature. To distinguish them, this section calls them scaling attention. Unlike dot-product attention which models global dependency, scaling attention reinforces the informative features and whittles the information-lacking features using a pooling layer [19]. Neither principles nor purposes of dot-product attention and scaling attention are completely divergent. In this Letter, we focus on dot-product attention.

### C. Semantic Segmentation

Fully Convolutional Network (FCN) based methods have witnessed enormous progress in semantic segmentation. DilatedFCN and EncoderDecoder are two prominent directions in FCN.

**DilatedFCN** The dilated or atrous convolution [20, 21] has shown its strong capability to retain the receptive field-of-view and enhance the performance of the backbone. Subsequent research concentrates on capture substantial contextual information, such as PSPNet [3] and DANet [4].

**EncoderDecoder** The EncoderDecoders utilize an encoder to capture multi-level feature maps, which are then incorporated into the final prediction by a decoder [1]. The refinement of the connection manners between the encoder and decoder [22] and the structure optimization based on residual connections are two significant directions [2].

## III. METHODOLOGY

In this section, we analyze the dot-product attention and formalize the proposed LAM. We demonstrate that substituting the attention from the conventional softmax function to the first-order approximation of Taylor expansion leads to linear time and memory complexity.

### A. Definition of Dot-Product Attention

Providing $N$ and $C$ as the length of input sequences and the number of input channels, where $N = H \times W$, and $H$ and $W$ denote the height and width of the input, with the input feature $X = [x_1, \cdots, x_N] \in \mathbb{R}^{N \times C}$, dot-product attention utilize three projected matrices $W_q \in \mathbb{R}^{D_x \times D_k}$, $W_k \in \mathbb{R}^{D_x \times D_k}$, and $W_v \in \mathbb{R}^{D_x \times D_v}$ to generate corresponding query matrix $Q$, the key matrix $K$, and the value matrix $V$:

$$Q = XW_q \in \mathbb{R}^{N \times D_k},$$

$$K = XW_k \in \mathbb{R}^{N \times D_k}, \qquad (1)$$

$$V = XW_v \in \mathbb{R}^{N \times D_v}.$$

Note the dimensions of the query matrix and key matrix have to be identical and all the vectors in this Letter are column vectors by default. A normalization function $\rho$ evaluates the similarity between the $i$-th query feature $q_i^T \in \mathbb{R}^{D_k}$ and the $j$-th key feature $k_j \in \mathbb{R}^{D_k}$ by $\rho(q_i^T k_j) \in \mathbb{R}^1$. Generally, as the query feature and key feature are generated by diverse layers, the similarities between $\rho(q_i^T k_j)$ and $\rho(q_j^T k_i)$ are not symmetric. By calculating the similarities between all pairs of positions and taking the similarities as weights, the dot-product attention module computes the value at position $i$ by aggregating the value features from all positions based on weighted summation:

$$D(Q, K, V) = \rho(QK^T)V. \quad (2)$$

The softmax is taken as the common normalization function:

$$\rho(Q^T K) = softmax_{row}(QK^T), \quad (3)$$

where $softmax_{row}$ indicates applying the softmax function along each row of matrix $QK^T$.

The $\rho(QK^T)$ models the similarities between each pair of pixels of the input, thoroughly extracting the global contextual information contained in the features. However, as $Q \in \mathbb{R}^{N \times D_k}$ and $K^T \in \mathbb{R}^{D_k \times N}$, the product between $Q$ and $K^T$ belongs to $\mathbb{R}^{N \times N}$, which leads to $O(N^2)$ memory and computational complexity. Therefore, the huge demand for resources in dot-product limits the application crucially on large inputs.

### B. Generalization of Dot-Product Attention

Under the condition of softmax normalization function, the $i$-th row of result matrix generated by the dot-product attention module according to equation (2) can be written as:

$$D(Q, K, V)_i = \frac{\sum_{j=1}^{N} e^{q_i^T k_j} v_j}{\sum_{j=1}^{N} e^{q_i^T k_j}}, \quad (4)$$

Equation (4) can then be generalized for any normalization function and rewritten as:

$$D(Q, K, V)_i = \frac{\sum_{j=1}^{N} sim(q_i, k_j) v_j}{\sum_{j=1}^{N} sim(q_i, k_j)}, \quad (5)$$
$$sim(q_i, k_j) \geq 0.$$

where $sim(q_i, k_j)$ can be expanded as $\phi(q_i)^T \varphi(k_j)$ which measures the similarity between the $q_i$ and $k_j$. Specifically, if $\phi(\cdot) = \varphi(\cdot) = e^{(\cdot)}$, equation (5) is equivalent to equation (4). As a consequence, equation (4) can be rewritten as:

$$D(Q, K, V)_i = \frac{\sum_{j=1}^{N} \phi(q_i)^T \varphi(k_j) v_j}{\sum_{j=1}^{N} \phi(q_i)^T \varphi(k_j)}, \quad (6)$$

and can be further simplified as:

$$D(Q, K, V)_i = \frac{\phi(q_i)^T \sum_{j=1}^{N} \varphi(k_j) v_j^T}{\phi(q_i)^T \sum_{j=1}^{N} \varphi(k_j)}. \quad (7)$$

As $K \in \mathbb{R}^{D_k \times N}$ and $V^T \in \mathbb{R}^{N \times D_v}$, the product between $K$ and $V^T$ belongs to $\mathbb{R}^{D_k \times D_v}$, which reduces the complexity of the dot-product attention mechanism considerably.

### C. Linear Attention Mechanism

Different from previous research, we conceive LAM based on the first-order approximation of Taylor expansion on equation (4):

$$e^{q_i^T k_j} \approx 1 + q_i^T k_j. \quad (8)$$

The above approximation, however, cannot guarantee the non-negativity. To ensure $q_i^T k_j \geq -1$, we can normalize $q_i$ and $k_j$ by $l_2$ norm:

$$sim(q_i, k_j) = 1 + \left(\frac{q_i}{\|q_i\|_2}\right)^T \left(\frac{k_j}{\|k_j\|_2}\right). \quad (9)$$

Then, equation (5) can be rewritten as:

$$D(Q, K, V)_i = \frac{\sum_{j=1}^{N} \left(1 + \left(\frac{q_i}{\|q_i\|_2}\right)^T \left(\frac{k_j}{\|k_j\|_2}\right)\right) v_j}{\sum_{j=1}^{N} \left(1 + \left(\frac{q_i}{\|q_i\|_2}\right)^T \left(\frac{k_j}{\|k_j\|_2}\right)\right)}, \quad (10)$$

and can be simplified as:

$$D(Q, K, V)_i = \frac{\sum_{j=1}^{N} v_j + \left(\frac{q_i}{\|q_i\|_2}\right)^T \sum_{j=1}^{N} \left(\frac{k_j}{\|k_j\|_2}\right) v_j^T}{N + \left(\frac{q_i}{\|q_i\|_2}\right)^T \sum_{j=1}^{N} \left(\frac{k_j}{\|k_j\|_2}\right)}. \quad (11)$$

The above equation can be transformed into a vectorized form as:

$$D(Q, K, V) = \frac{\sum_j V_{i,j} + \left(\frac{Q}{\|Q\|_2}\right)\left(\left(\frac{K}{\|K\|_2}\right)^T V\right)}{N + \left(\frac{Q}{\|Q\|_2}\right)\sum_j \left(\frac{K}{\|K\|_2}\right)^T_{i,j}}. \quad (12)$$

As $\sum_{j=1}^{N}\left(\frac{k_j}{\|k_j\|_2}\right) v_j^T$ and $\sum_{j=1}^{N}\left(\frac{k_j}{\|k_j\|_2}\right)$ can be calculated and reused for every query, time and memory complexity of the proposed LAM based on equation (12) is $O(N)$.

### D. Multi-stage Attention ResU-Net

We design the attention block based on the proposed LAM to capture global contextual information, as shown in Fig 1(b). Considering the channels of the input $C$ is normally far less than the number of pixels (i.e., $C \ll N$), the complexity of the softmax function for channels (i.e., $O(C^2)$), is not high based on equation (3). Thus, for the channel dimension, we directly harness dot-product-based attention.

As the complexity is dramatically decreased by proposed LAM, the usage of the attention block for large feature maps is brought within the bounds of possibility. Thus, by taking the ResNet as the backbone, we design the MAResU-Net to combine the low-level and high-level feature maps through attention block in multiple stages, as shown in Fig 1(a). To design a concise but effective framework, we select the ResNet-18 and ResNet-34 as backbones rather than more complicated ResNets (e.g. ResNet-101).

TABLE I
THE EXPERIMENTAL RESULTS ON VAIHINGEN DATASETS.

| Method | Backbone | Imp. surf. | Building | Low veg. | Tree | Car | Mean F1 | OA (%) | mIoU (%) |
|---|---|---|---|---|---|---|---|---|---|
| U-Net [1] | - | 84.331 | 86.479 | 73.132 | 83.886 | 40.825 | 73.731 | 82.023 | 61.362 |
| ResUNet-a [2] | - | 86.708 | 88.319 | 76.791 | 85.429 | 57.094 | 78.868 | 84.350 | 66.995 |
| PSPNet [3] | ResNet34 | 90.273 | 94.218 | 82.757 | 88.606 | 51.100 | 81.391 | 88.820 | 71.591 |
| DANet [4] | ResNet34 | 91.135 | 94.818 | 83.467 | 88.923 | 62.979 | 84.264 | 89.524 | 74.728 |
| EaNet [7] | ResNet34 | 92.172 | 95.197 | 82.811 | 89.254 | 80.563 | 87.999 | 89.995 | 80.223 |
| CE-Net [9] | ResNet34 | 92.681 | **95.529** | 83.359 | 89.492 | 81.243 | 88.461 | 90.402 | 81.492 |
| MAResU-Net | ResNet18 | 91.971 | 95.044 | 83.735 | 89.349 | 78.283 | 87.676 | 90.047 | 80.749 |
| MAResU-Net | ResNet34 | **92.912** | 95.256 | **84.947** | **89.939** | **88.330** | **90.277** | **90.860** | **83.301** |

TABLE II
KAPPA Z-TEST COMPARING THE PERFORMANCE OF DIFFERENT METHODS.

| Method | Kappa | KV ($10^{-9}$) | ResUNet-a | PSPNet | DANet | EANet | CE-Net | MLAU-Net |
|---|---|---|---|---|---|---|---|---|
| U-Net | 0.7682 | 3.1443 | 12.7608 | 39.5863 | 43.8255 | 47.5193 | 50.3952 | 52.8544 |
| ResUNet-a | 0.7993 | 2.7954 | - | 26.8824 | 31.1415 | 34.8537 | 37.7492 | 40.2256 |
| PSPNet | 0.8586 | 2.0706 | - | - | 4.2844 | 8.0201 | 10.9479 | 13.4527 |
| DANet | 0.8672 | 1.9586 | - | - | - | 3.7358 | 6.6662 | 9.1734 |
| EaNet | 0.8745 | 1.8598 | - | - | - | - | 2.9328 | 5.4420 |
| CE-Net | 0.8801 | 1.7861 | - | - | - | - | - | 2.5092 |
| MAResU-Net | 0.8848 | 1.7224 | - | - | - | - | - | - |

TABLE III
THE ABLATION STUDY ABOUT THE ATTENTION BLOCK.

| Method | Mean F1 | OA (%) | mIoU |
|---|---|---|---|
| ResNet34 | 85.897 | 89.495 | 78.520 |
| MAResU-Net 1 | 86.159 | 89.543 | 79.420 |
| MAResU-Net 2 | 87.397 | 89.794 | 80.009 |
| MAResU-Net 3 | 88.952 | 90.025 | 81.184 |
| MAResU-Net 4 | 89.492 | 90.468 | 82.101 |
| MAResU-Net | 90.277 | 90.860 | 83.301 |

TABLE IV
THE TRAINING TIME, COMPLEXITY AND PARAMETERS.

| Method | Time (s) / epoch | Complexity (G) | Parameters (M) |
|---|---|---|---|
| U-Net | 251 | 61.921 | 43.420 |
| ResUNet-a | 511 | 26.991 | 39.946 |
| PSPNet | 101 | 5.563 | 34.138 |
| DANet | 85 | 4.890 | 22.782 |
| EaNet | 108 | 7.107 | 44.341 |
| CE-Net | 124 | 8.988 | 29.005 |
| MAResU-Net | 135 | 8.772 | 26.277 |

## IV. EXPERIMENTAL RESULTS

### A. Dataset

The Vaihingen semantic labeling dataset [23] is composed of 33 tiles, with an average size of 2494 × 2064 pixels and a GSD of 5 cm. The near-infrared, red, and green channels together with DSM are provided in the dataset. There are 16 image tiles in the training set and 17 image tiles in the test set. Concretely, we exploit ID: 2, 4, 6, 8, 10, 12, 14, 16, 20, 22, 24, 27, 29, 31, 33, 35, 38 for testing, ID: 30 for validation, and the remaining 15 images for training. Please note that we do not use DSM in our experiments for simplicity.

### B. Experimental Setting

All of the experiments are implemented with PyTorch on a single Tesla V100, and the optimizer is set as AdamW with a 0.0003 learning rate. The soft cross-entropy loss function is used as a quantitative evaluation, and a backpropagation index is adopted to measure the disparity between the obtained 2D segmentation maps and ground truth. For each method, the overall accuracy (OA), mean Intersection over Union (mIoU), and F1-score (F1) are chosen as evaluation indexes. OA is computed for all categories including the background.

### C. Semantic Segmentation Results and Analysis

*1) Performance Comparison:* The experimental results of different methods on the Vaihingen dataset are demonstrated in Table I. The proposed MAResU-Net based on ResNet-34 achieves the highest mean F1-score of 90.277 %, OA of 90.860%, and mIoU of 83.301%, which is higher than recent contextual information aggregation methods designed initially for natural images, such as PSPNet and DANet, and also prevail over lastest multi-scale feature aggregation models proposed for remote sensing images, such as ResUNet-a and EaNet.

The computational demand for attention in memory and computation cost has significantly reduced thanks to LAM. Therefore, it is possible to directly model the global contextual dependencies of each pair of positions regardless of the size of the input, thereby enhancing the performance of the network in multiple stages. The superiority of the proposed MAResU-Net demonstrates that our method can capture refined and fine-grained features. The visual comparison is shown in Fig. 2 with the clear advantages in our method compared with other benchmarks.

*2) Statistical Significance:* To evaluate the statistical significance, we report Kappa z-test for pairwise methods based on Kappa coefficients of agreement and their variances using the following equation:

$$z = (k_1 - k_2)/\sqrt{v_1 + v_2}. \quad (13)$$

where $k$ indicates the Kappa coefficient and $v$ represents the Kappa variance. Specifically, if the value of $z$ surpasses a threshold of 1.96, the two methods are regarded signally different at the 95 % confidence level. As can be seen from Table II, the classification accuracy of the proposed MAResU-Net is statistically higher than other comparative methods.

*3) Ablation Study:* As multi-stage attention blocks are utilized to exploit global contextual in the proposed MAResU-Net, it is valuable to conduct the ablation study and measure the impact of each block. As shown in TABLE III, the utilization of attention blocks boosts the classification performance dramatically compared with the baseline methods. Meanwhile, attention blocks at low levels contribute more than those at high levels.

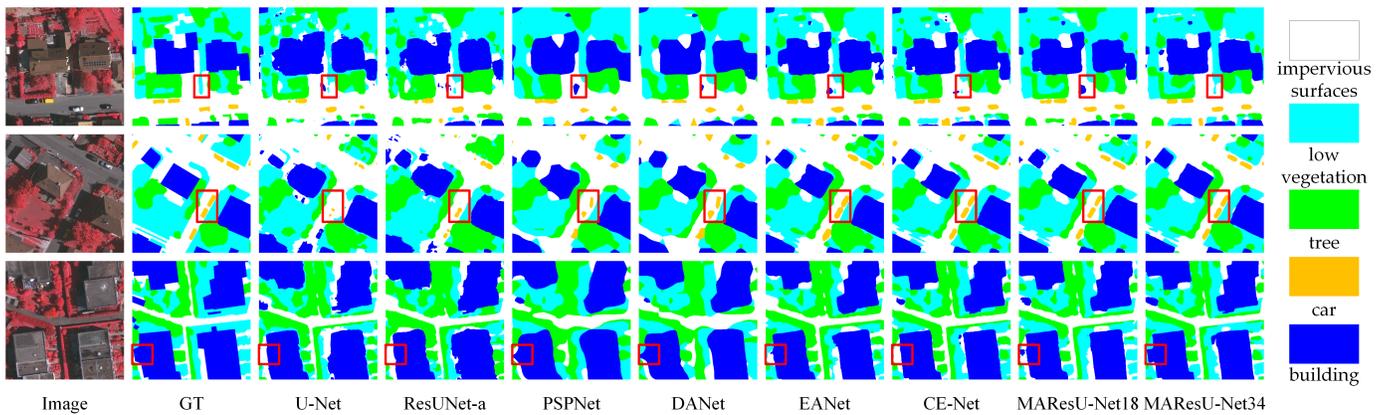

Fig. 2. The overall architecture of proposed MAResU-Net in (a), and the structure of the attention block in (b).

*4) Model Complexity:* Considering the complexity of the model is significant to assess the merit of a framework, we report the training time for each epoch, complexity, and parameters in Table IV, which demonstrates that the design of MAResU-Net is computationally efficient.

## V. CONCLUSION

In this Letter, we propose a linear attention mechanism which reduces the complexity of the dot-product attention mechanism from $O(N^2)$ to $O(N)$. Based on the proposed LAM and the ResNet, we reconstruct the skip connections in the raw U-Net and proposed the MAResU-Net for semantic segmentation using fine-resolution remote sensing images. Experiments conducted on the Vaihingen dataset manifest the effectiveness of our MAResU-Net in both classification accuracy and computational efficiency.

## ACKNOWLEDGMENTS

The design of the method was inspired by Jianlin Su's blog in https://spaces.ac.cn/archives/7546 (04 Jul 2020).

## REFERENCE

[1] O. Ronneberger, P. Fischer, and T. Brox, "U-net: Convolutional networks for biomedical image segmentation," in *International Conference on Medical image computing and computer-assisted intervention*, 2015: Springer, pp. 234-241.

[2] F. I. Diakogiannis, F. Waldner, P. Caccetta, and C. Wu, "Resunet-a: a deep learning framework for semantic segmentation of remotely sensed data," *ISPRS J. Photogramm. Remote Sens.,* vol. 162, pp. 94-114, 2020.

[3] H. Zhao, J. Shi, X. Qi, X. Wang, and J. Jia, "Pyramid scene parsing network," in *Proceedings of the IEEE conference on computer vision and pattern recognition*, 2017, pp. 2881-2890.

[4] J. Fu *et al.*, "Dual attention network for scene segmentation," in *Proceedings of the IEEE Conference on Computer Vision and Pattern Recognition*, 2019, pp. 3146-3154.

[5] J. Devlin, M.-W. Chang, K. Lee, and K. Toutanova, "Bert: Pre-training of deep bidirectional transformers for language understanding," *arXiv preprint arXiv:1810.04805,* 2018.

[6] A. Vaswani *et al.*, "Attention is all you need," in *Advances in neural information processing systems*, 2017, pp. 5998-6008.

[7] X. Zheng, L. Huan, G.-S. Xia, and J. Gong, "Parsing very high resolution urban scene images by learning deep ConvNets with edge-aware loss," *ISPRS J. Photogramm. Remote Sens.,* vol. 170, pp. 15-28, 2020.

[8] X. Wang, R. Girshick, A. Gupta, and K. He, "Non-local neural networks," in *Proceedings of the IEEE conference on computer vision and pattern recognition*, 2018, pp. 7794-7803.

[9] Z. Gu *et al.*, "Ce-net: Context encoder network for 2d medical image segmentation," *IEEE transactions on medical imaging,* vol. 38, no. 10, pp. 2281-2292, 2019.

[10] R. Child, S. Gray, A. Radford, and I. Sutskever, "Generating long sequences with sparse transformers," *arXiv preprint arXiv:1904.10509,* 2019.

[11] N. Kitaev, Ł. Kaiser, and A. Levskaya, "Reformer: The efficient transformer," *arXiv preprint arXiv:2001.04451,* 2020.

[12] A. Katharopoulos, A. Vyas, N. Pappas, and F. Fleuret, "Transformers are RNNs: Fast Autoregressive Transformers with Linear Attention," *arXiv preprint arXiv:2006.16236,* 2020.

[13] R. Li, S. Zheng, C. Duan, and J. Su, "Multi-Attention-Network for Semantic Segmentation of High-Resolution Remote Sensing Images," *arXiv preprint arXiv:2009.02130,* 2020.

[14] Z. Shen, M. Zhang, H. Zhao, S. Yi, and H. Li, "Efficient Attention: Attention with Linear Complexities," *arXiv preprint arXiv:1812.01243,* 2018.

[15] Z. Zhou, M. M. R. Siddiquee, N. Tajbakhsh, and J. Liang, "Unet++: A nested u-net architecture for medical image segmentation," in *Deep Learning in Medical Image Analysis and Multimodal Learning for Clinical Decision Support*: Springer, 2018, pp. 3-11.

[16] K. Yue, L. Yang, R. Li, W. Hu, F. Zhang, and W. Li, "TreeUNet: Adaptive Tree convolutional neural networks for subdecimeter aerial image segmentation," *ISPRS J. Photogramm. Remote Sens.,* vol. 156, pp. 1-13, 2019.

[17] C. Zhang, P. M. Atkinson, C. George, Z. Wen, M. Diazgranados, and F. Gerard, "Identifying and mapping individual plants in a highly diverse high-elevation ecosystem using UAV imagery and deep learning," *ISPRS J. Photogramm. Remote Sens.,* vol. 169, pp. 280-291, 2020.

[18] D. Bahdanau, K. Cho, and Y. Bengio, "Neural machine translation by jointly learning to align and translate," *arXiv preprint arXiv:1409.0473,* 2014.

[19] J. Hu, L. Shen, and G. Sun, "Squeeze-and-excitation networks," in *Proceedings of the IEEE conference on computer vision and pattern recognition*, 2018, pp. 7132-7141.

[20] F. Yu and V. Koltun, "Multi-scale context aggregation by dilated convolutions," *arXiv preprint arXiv:1511.07122,* 2015.

[21] L.-C. Chen, G. Papandreou, I. Kokkinos, K. Murphy, and A. L. Yuille, "Semantic image segmentation with deep convolutional nets and fully connected crfs," *arXiv preprint arXiv:1412.7062,* 2014.

[22] L. Rui, D. Cehnxi, and Z. Shunyi, "MACU-Net Semantic Segmentation from High-Resolution Remote Sensing Images," *arXiv preprint arXiv:2007.13083,* 2020.

[23] "ISPRS 2D semantic labeling contest." [Online]. Available: http://www2.isprs.org/commissions/comm3/wg4/semantic-labeling.html.